\pdfoutput=1

\documentclass[11pt]{article}

\usepackage{acl}

\usepackage{times}
\usepackage{latexsym}
\usepackage{tabularx}
\usepackage{makecell}
\usepackage{graphicx}
\usepackage{multirow}
\usepackage{color,soul}
\usepackage{float}

\usepackage[T1]{fontenc}
\usepackage[utf8]{inputenc}

\usepackage{microtype}
\usepackage{inconsolata}
\usepackage[linesnumbered,ruled,vlined]{algorithm2e}
\usepackage{xcolor} %
\usepackage{tikz}
\usetikzlibrary{fit,calc}

\colorlet{mypink}{red!40}
\colorlet{myblue}{cyan!60}

\title{Leveraging Interesting Facts to Enhance User Engagement with Conversational Interfaces}

\author{Nikhita Vedula ~~~ Giuseppe Castellucci ~~~ Eugene Agichtein \\
        \textbf{Oleg Rokhlenko \and Shervin Malmasi} \\
         Amazon.com, Inc. ~~~ Seattle, WA, USA \\
         \texttt{\{veduln,giusecas,eugeneag,olegro,malmasi\}@amazon.com}}

\begin{document}
\maketitle
\begin{abstract}

Conversational Task Assistants (CTAs) guide users in performing a multitude of activities, such as making recipes. %
However, ensuring that interactions remain engaging, interesting, and enjoyable for CTA users is not trivial, especially for time-consuming or challenging tasks.
Grounded in psychological theories of \textit{human interest}, we propose to engage users with contextual and interesting statements or \textit{facts} during interactions with a multi-modal CTA, to reduce fatigue and task abandonment before a task is complete. 
To operationalize this idea, we train a high-performing classifier ($82\%$ F1-score) to automatically identify relevant and interesting facts for users. We use it to create an annotated dataset of \textit{task-specific interesting facts}\footnote{\label{note1}\url{www.github.com/vnik18/cta-interesting-facts}} for the domain of cooking. Finally, we design and validate a dialogue policy to incorporate the identified relevant and interesting facts into a conversation, to improve user engagement and task completion. 
Live testing on a leading multi-modal voice assistant shows that $66\%$ of the presented facts were received positively, leading to a $40\%$ gain in the user satisfaction rating, and a $37\%$ increase in conversation length. 
These findings emphasize that strategically incorporating interesting facts into the CTA experience can promote real-world user participation for guided task interactions. 

\end{abstract}

\section{Introduction}
\label{sec:intro}

Conversational Task Assistants (CTAs) are a class of conversational agents that guide human users step-by-step in performing a multitude of activities, like cooking or home improvement tasks~\cite{vtyurina2018exploring,strathearn2022task2dial,choi2022wizard,he2023understanding,agichtein2023advancing}.
CTAs help users search and find the right task for their needs, inform about the task execution (e.g., tools to use), and answer questions about the task (e.g., substituting an ingredient).

\begin{figure}[t]
\centering
\includegraphics[width=0.44\textwidth]{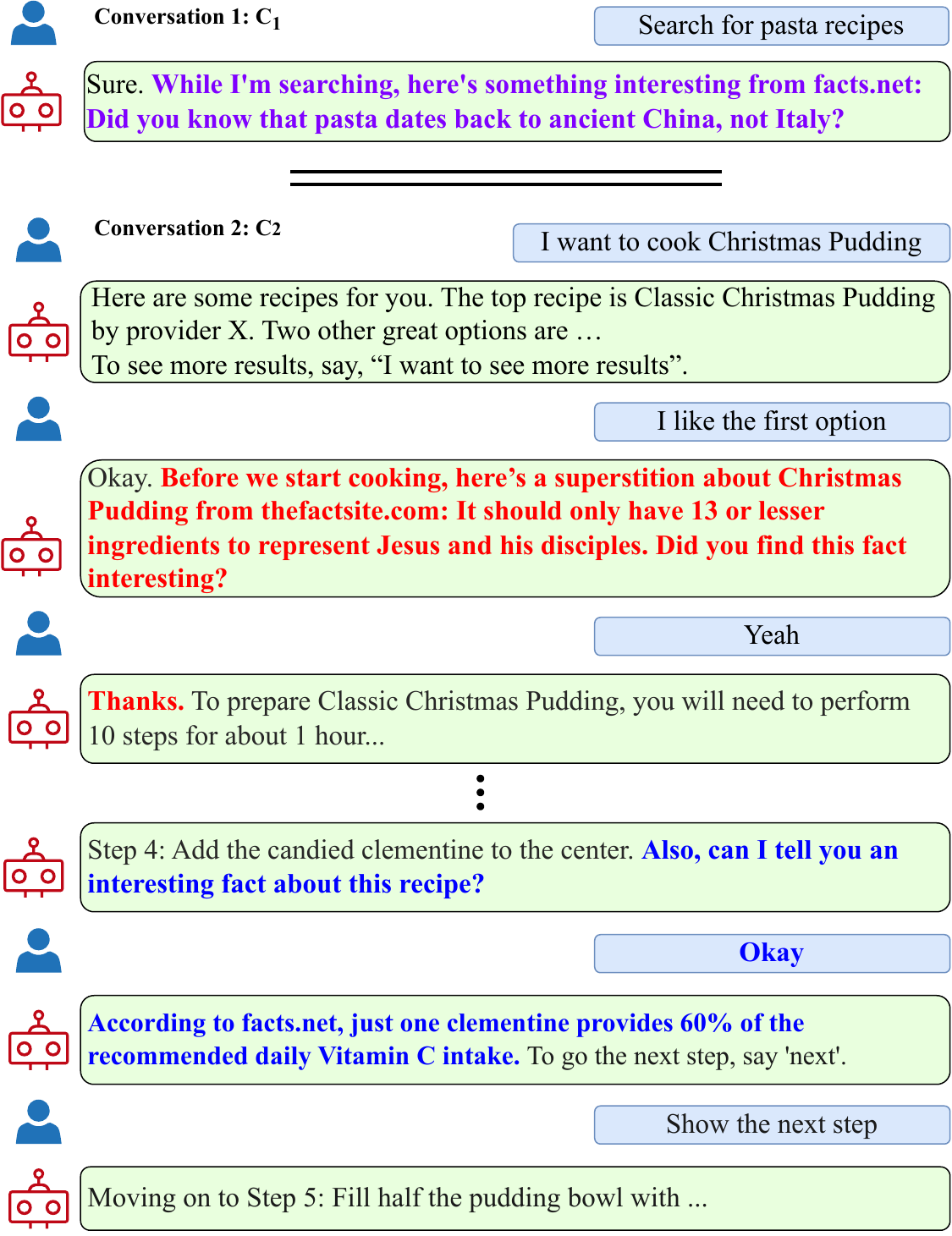}
\caption{Conversations $C_1$ and $C_2$ between two users and our CTA. $C_1$ shows an interesting fact during search. $C_2$ has two interesting facts during task execution.}
\label{fig1}
\end{figure}

Besides ensuring that CTAs provide useful, accurate content, %
it is crucial to encourage users to consistently use CTAs to accomplish long and complex tasks.
In this setting, users can easily get bored and abandon tasks, so maintaining interesting and entertaining conversations is key \cite{folstad2020users,ceha2021can}. %
Understanding and predicting human interest is crucial to engage and retain users \cite{constantin2019computational}.

We hypothesize that incorporating relevant, entertaining pieces of information into conversations is a powerful tool to maintain and sustain user attention, satisfaction, and participation.
For example, while cooking with a CTA, users are likely to interact with it longer, frequently and repeatedly, if their experience is enriched with intriguing facts about the recipe's history, nutritional benefits of its ingredients, or useful tips about its cooking techniques (see Figure~\ref{fig1} for an example).
Such information can increase user knowledge, satisfaction and trust in the CTA by establishing it as a knowledgeable domain expert. It transforms users' dialogue with the CTA into more than just task-oriented, resulting in an improved user experience, particularly during long, tedious, or monotonous activities. 

We propose a practical approach to enhance user engagement with CTAs. 
Grounded in socio-psychological theories on human \textit{interest}~\cite{berlyne1949interest,berlyne1970novelty,silvia2005interesting}, we design a novel schema to comprehensively and objectively examine the \textit{interestingness} of pieces of information (henceforth called \textit{`facts'}) in a CTA setting. 
We study if such facts can spark and foster user interest, and thus boost user satisfaction and engagement during a conversation with a CTA.
We also explore multiple techniques to share interesting facts with users. %
We perform live testing of our framework in a leading multi-modal voice assistant. %
$66\%$ of our facts were received positively by real-world users, resulting in a $40\%$ gain in average user satisfaction ratings, and a $37\%$ increase in average conversation length. 
These results demonstrate the practical benefits of enriching the CTA experience with interesting facts, and showcase our approach's potential to enhance the adoption of CTAs for complex real-world tasks requiring sustained user engagement.

We also create for public release a dataset of $1,379$ interesting facts associated with cooking, extracted from online sources and cleaned for grammar, style, relevance, interestingness, and appropriateness. We believe that this can be a valuable resource to drive future research in CTAs, and user engagement in task-assistance conversations. %

\section{Related Work}
\label{sec:related}

\paragraph{Interest and Interestingness:}
Our work on enhancing user engagement with CTAs by increasing their level of interest in the conversation is inspired by multiple psychological theories conceptualizing the human behavioral dimensions of interest and interestingness. 
One of the earliest theories is that of Berlyne~\cite{berlyne1949interest,berlyne1970novelty}, who viewed interest as a defining factor of motivation, influenced by several factors including uncertainty, complexity and novelty of the subject at hand. ~\citet{tomkins1962affect,silvia2005interesting} defined interest as an appealing effect of an activity on a human caused by novelty, exploration, learning, creativity and affect, leading to increased interaction with the subject matter. 

\citet{geng2006interestingness} surveyed multiple features relevant to interestingness such as conciseness, coverage, reliability, novelty, surprise and utility in the context of pattern mining and association rules. \citet{carvalho2005evaluating} studied their correlation with real human interest.
\citet{wu2013curiosity,constantin2019computational} observed a negative correlation between interest and low-arousal emotions like memorability, familiarity, and boredom. 
\citet{clark2019makes,ceha2021can} employed humor, jokes and fun facts within their conversation agents to demonstrate that providing such entertaining information to users can increase user interest, engagement and retention, and reduce monotony.  

In our work, we model several aspects contributing to human interest in Section~\ref{sec:methods}, to identify facts that can interest and engage users interacting with a CTA, and reduce monotony. We show in Section~\ref{sec:experiments} that a CTA without interest enhancing signals can cause boredom and reduce engagement by users.

\paragraph{Conversational Task Assistants:}
CTAs are a special type of conversational assistants~\cite{agichtein2023advancing}, different from general purpose intelligent assistants (e.g., Amazon's Alexa, Apple's Siri). CTAs guide users to perform tasks in a step-wise manner, also answering user questions along the way. CTAs differ from conversational search agents~\cite{vtyurina2017exploring} and task-oriented conversational agents~\cite{zhang2020recent,strathearn2022task2dial}, where the agent and not the user performs a task based on user input (e.g., booking a ticket). 
User satisfaction and engagement has been widely studied in the areas of search, task-oriented dialog systems and virtual assistants~\cite{kiseleva2016predicting,park2020large,sekulic2021user,lotze2021ranking,siro2022understanding}. However, these approaches lack a theoretical backing and have not been studied for CTAs. 

The work that bears the most similarity with ours is Wizard of Curiosities (WoC)~\cite{vicente2023wizard}, which also adds facts %
within dialogues between users and CTAs. %
However, both works differ in the following ways: (i) %
WoC targets \textit{information-seeking} traits of humans based on \textit{curiosity} theory~\cite{berlyne1966curiosity,kidd2015psychology}. %
Our work is more principled and draws upon the broader theories of \textit{interest} which encompass informativeness and curiosity, and are better suited for studying and improving user engagement~\cite{pekrun2019murky}; %
(ii) WoC does not consider several features that we consider in this work to add facts into the CTA dialogue flow, namely novelty, recency, frequency, interest and emotion (defined in Section~\ref{sec:methods}). 
(iii) Unlike WoC, we translate the theoretical aspects into a novel annotation schema for interestingness that is more objective, consistent and precise for a CTA (Sections~\ref{sec:schema} and~\ref{sec:design}). 
(iv) We study several user experience options on how to effectively present users interesting facts during the conversation flow.

\section{Boosting CTA Engagement with Facts}
\label{sec:methods}

We now describe our framework to offer interesting facts to users conversing with a multi-modal CTA, having both display and voice components, to entertain users and increase their engagement. 
We focus on cooking-specific conversations, but as explained in Section~\ref{sec:discussion}, we believe that our framework can generalize to any CTA-feasible domain.

\subsection{Feature Schema for Interestingness}
\label{sec:schema}

We first define a feature schema inspired by socio-psychological work on interestingness~\cite{berlyne1949interest,berlyne1970novelty}, and use it to curate a dataset of facts that may increase user interest and lower monotony.

\textbf{(i) Conciseness:} A fact should not be too long in terms of number of words and listening time.

\textbf{(ii) Specificity:} A fact should be understandable in general to a large number of users, without requiring specific background or domain knowledge. 

\textbf{(iii) Novelty:} A fact should offer new, unique information that may surprise an average user not a knowledge expert in the domain.  

\textbf{(iv) Relevance:} A fact should be relevant and related to the considered conversation turn. 

\textbf{(v) Informativeness:} A fact should deliver useful, helpful knowledge or insights about the task.

\subsection{Interesting Facts Dataset}
\label{sec:factdataset}

We create a dataset of $1,379$ interesting facts associated with cooking, following the above feature schema.
We use an off-the-shelf named entity recognition model to extract a list $\mathcal{E}$ of entities of the types \textit{ingredient}, \textit{recipe}, and \textit{tool}, from a large corpus of recipes $\mathcal{R}$.\footnote{The recipe provider in this work is Whole Foods Recipes.} We then crawl specific websites\footnote{Wikipedia, and fun facts websites like \url{www.facts.net}} to extract diverse candidate facts for $\mathcal{E}$.

To filter out irrelevant facts, we split each fact into individual sentences. We then fine-tune and use a domain classifier, consisting of a linear layer on top of the 5B Alexa Teacher Model~\cite{fitzgerald2022alexa}, with positive training sentences from $\mathcal{R}$, and negative instances from out-of-domain online sources.\footnote{The model achieves a $0.91$ F1-score on a 13.8K test set.}
We match each of the domain-relevant candidate facts to our list of identified entities $\mathcal{E}$. We only retain facts 
if they contain an entity in $\mathcal{E}$ as either the subject or the object of the fact sentence, based on its dependency parse tree.\footnote{We use \url{www.spacy.io} to compute the parse tree.} This gives us a set of potential interesting facts $\mathcal{F'}$.

We next sample a set of $750$ facts from $\mathcal{F'}$, to be manually annotated for their level of \textit{interestingness}. For each fact, $2$ expert annotators familiar with this domain give it a binary label with respect to each feature of our schema (Section~\ref{sec:schema}), followed by an overall label for interestingness with respect to the entity under consideration. %
We also ask annotators to choose the feature most important to them to decide if a fact selected by a CTA could be interesting. Having such a structured schema can mitigate the subjectivity and/or biases likely to occur while annotating the inherently subjective element of \textit{interestingness}, giving more objective, consistent, and precise annotations. 

We find that annotators prefer novel and easily understood facts. 
Aggregating their preferences yields this ranked order of feature importance for relevant facts: novelty, specificity, conciseness and informativeness, with a per-feature binary inter-annotator agreement of $0.68$, $0.84$, $0.88$ and $0.75$ respectively. We assign each feature a normalized weight based on how many times they were chosen as important. We then compute a linearly combined interestingness score for each fact, resulting in a set $\mathcal{F}$ of $606$ annotated interesting facts. These facts serve as positive instances, and irrelevant or uninteresting facts from $\mathcal{F'}$ serve as negative instances, to train a multi-label binary classifier on top of a RoBERTa~\cite{liu2019roberta} language model. The two labels denote relevance and interestingness to the input text (the current step of the ongoing task).

We use this classifier to automatically label the remaining domain-relevant candidate facts in $\mathcal{F'}$. Each fact is linked to a step in the corpus $R$, based on the entities present in the task step.  
To minimize redundancy and ensure that the facts associated with each step are sufficiently diverse, we cluster and filter facts with a high ($\geq0.85$) pairwise cosine similarity based on Sentence-BERT~\cite{reimers2019sentence} embeddings.
Each fact is then manually cleaned for grammar, style and appropriateness. We finally obtain a novel dataset of $1,379$ interesting cooking facts (see dataset details in Table~\ref{tab:stats}), to present to real-time customers conversing and performing a task with a CTA. Each fact has accompanying evidence in the form of the website URL it is sourced from.

\begin{algorithm}
  \caption{Presenting interesting facts to a user interacting with a CTA during search.}
  \label{algorithm:1}
  
  \KwData{Facts associated with each task}
  
   \hspace{-2pt} $max\_facts \gets initialize$; $min\_turns\_btw\_facts \gets initialize$\;
  
  \SetAlgoNlRelativeSize{-1}

  \SetKwFunction{showFactAtStep}{showFactAtStep}
  
  \SetKwProg{Fn}{Function}{:}{}

  \Fn{\showFactAtStep{}}{
    \tcp{Show user fact at this step?}
    \If {fact exists for step \textbf{and}
    \# facts already shown < $max\_facts$ \textbf{and}
    \# turns betw. facts $\geq$ $min\_turns\_btw\_facts$ \textbf{and} final turn is voice-friendly}
    {
    \textbf{return} \textit{True}\;
  }
  \Else{
    \textbf{return} \textit{False}\;
  }
  }
  
  User searches for a task

  \If{\showFactAtStep{}}{
    Present current search result info\;
    Show interesting fact \& record count\;
  }

  Repeat steps 8-10 until the user chooses a task to execute \textbf{or} exits conversation\;

\end{algorithm}

\subsection{Incorporating Interesting Facts into Conversations}    
\label{sec:design}

The interestingness feature schema described in Section~\ref{sec:schema} only includes \textit{fact-specific} features, to recognize a given fact as interesting.  
Recall that a CTA aids users explore or search for tasks, provides information about tasks, and guides users through step-by-step execution instructions.
For a satisfying user experience (UX), several \textit{conversational} dimensions need to be considered to effectively inject interesting facts into user-CTA interactions. We consider alternative dialogue policies:

(i) \textbf{Present Facts Before \& During a Task}: We offer interesting facts to users both when they are exploring or searching for a task to perform, and during the step-by-step execution of their chosen task. Algorithms~\ref{algorithm:1} and~\ref{algorithm:2} show both these UXs. %

(ii) \textbf{Select Turns to Present Facts:} %
Engaging users is crucial, but the goal of a user-CTA interaction is to complete a task (e.g., cook a recipe). We use the below criteria to avoid creating distractions and a poor experience by showing too many facts. 

\begin{algorithm}
  \caption{Presenting interesting facts while a user executes a task with a CTA.}
  \label{algorithm:2}
  
  \KwData{Facts associated with chosen task}
  
   \hspace{-2pt} $max\_facts \gets initialize$; $min\_turns\_btw\_facts \gets initialize$\;
  
  \SetAlgoNlRelativeSize{-1}
  \SetKwFunction{seekUserFeedback}{seekUserFeedback}
  
  \SetKwProg{Fn}{Function}{:}{}

  \Fn{\seekUserFeedback{}}{
    \tcp{Seek user feedback about fact}
    \If{fact shown \normalfont{\textbf{and}} \textit{feedback not sought before}}{
    Ask user if fact was interesting\;
    Record user response\;
  }
  }

  Step-by-step task execution

  \If{\showFactAtStep{}}{
    Present current step text\;
    Ask user if they want to see a fact\;
    If user agrees, show fact \& record count\;
    If user rejects, set $max\_facts = 1$\;
  }

  \seekUserFeedback{} about fact shown

  Repeat steps 6-12 until user completes the task execution \textbf{or} exits the conversation\;

\end{algorithm}

\begin{itemize}

\item \textbf{Recency:} %
After presenting a fact, the next one is shown only after a set number of turns.

\item \textbf{Frequency}: A CTA should not present too many facts per conversation. %
The recency and frequency parameters are implemented in the \texttt{showFactAtStep()} function in Algorithms~\ref{algorithm:1} and~\ref{algorithm:2}, and their values are chosen empirically.

\item \textbf{Diversity}: Our dataset creation in Section~\ref{sec:factdataset} ensures that users are shown diverse facts. %

\item \textbf{Voice-friendly}: The word count %
of all conversation turns with interesting facts is bounded. %

\end{itemize}

(iii) \textbf{Proactive Inquiry}: A CTA explicitly asking a user if they want to see an interesting fact or not adds two extra turns in the conversation, but 
reduces the possibility of distracting or frustrating users with undesirable information. 
When we present facts to users during their search (step 10 of Algorithm~\ref{algorithm:1}, first two turns of Figure~\ref{fig1}), we do not seek prior permission from them, to avoid additional dialogue overhead before the task even begins. 
However, during task execution, our CTA seeks user approval each time before showing interesting facts (steps 9-10 of Algorithm~\ref{algorithm:2}). Users can choose to not answer or bypass this request, in which case the next step of the task continues. %

(iv) \textbf{Seek User Feedback on Facts:} We explicitly ask the user \textit{once} during the conversation if they liked the fact that was shown to them (\texttt{seekUserFeedback()} function in Algorithm~\ref{algorithm:2}). Users can choose to not answer or bypass this question, and continue to execute their task normally.

\section{Experiments and Results}
\label{sec:experiments}

\subsection{Interesting Fact Quality Evaluation}
\label{sec:automatic-eval}

\begin{table}[t]
\footnotesize
\centering
\begin{tabular}{l|r}
\hline
Number of interesting facts     & 1,379   \\ %
Number of unique fact providers    &  5  \\ %
Number of unique entities & 420 \\
Mean fact length in words    &  13   \\ \hline
\end{tabular}
\caption{Details of our curated interesting facts dataset.}
\label{tab:stats}
\end{table}

\begin{table}[t]
\centering
\resizebox{1\columnwidth}{!}{
\begin{tabular}{l|c|c|c}
\hline
\textbf{Model} & \textbf{F1-relevance} & \textbf{F1-interestingness} & \textbf{F1-rel-int} \\
\hline
\texttt{roberta-base-rel}     & 0.76 & N/A & 0.76  \\ %
\texttt{roberta-base-int}  & N/A & 0.67 & 0.67  \\ %
\texttt{roberta-base-rel-int} (Ours)  & 0.83 & \textbf{0.80} & \textbf{0.82}  \\ %
\texttt{gpt-3.5-turbo-fs-rel}  & \textbf{0.84} & N/A & N/A  \\ %
\texttt{gpt-3.5-turbo-fs-int}  & N/A & 0.72 & N/A  \\ %
\texttt{gpt-3.5-turbo-fs-rel-int}  & 0.78 & 0.66 & 0.72  \\ \hline
\end{tabular}}
\caption{Detecting relevance and interestingness of facts. `rel', `int' and `fs' denote relevance, interestingness, and few shot respectively. N/A means that a model does not predict the corresponding label.}
\label{tab:classifier}
\end{table}

\paragraph{Relevance and Interestingness Detection:}
In Section~\ref{sec:factdataset}, we proposed a multi-label classifier to detect relevant and interesting facts. We now assess multiple approaches in detecting relevance and interestingness in Table~\ref{tab:classifier}, on a test set of $400$ facts with a $50\%$ split of positive and negative candidates. Our proposed classifier (\texttt{roberta-base-rel-int}) improves upon the single-label classifier baselines (\texttt{roberta-base-rel, roberta-base-int}) by at least $5\%$ F1 score points, in detecting if an input fact is relevant and interesting. 
We also prompt the \texttt{gpt-3.5-turbo}~\cite{open2023introducing} LLM with $8$ randomly chosen few-shot examples from our training set to recognize relevance and interestingness individually (\texttt{gpt-3.5-turbo-fs-rel} and \texttt{gpt-3.5-turbo-fs-int}), as well as together in the same prompt (last row of the table). %
Our multi-label classifier is comparable to GPT-3.5 in detecting relevant facts despite being a much smaller model. Moreover, it significantly outperforms GPT-3.5 by $10\%$ in detecting the interestingness of facts.

\paragraph{Evaluating LLMs for Annotation:}

We prompt GPT-4~\cite{openai2023gpt} %
with our well-defined interestingness feature schema from Section~\ref{sec:schema}. Similar to the human annotation in Section~\ref{sec:factdataset}, we ask GPT-4 to label (i) each input fact for specificity, novelty, relevance and informativeness; (ii) overall interestingness based only on the schema features; and (iii) the most important feature to decide a fact's interestingness.
On $75$ randomly selected facts from Section~\ref{sec:factdataset}, we observe $44\%$ agreement with human labels for overall interestingness, showing that it is non-trivial to replace humans with LLMs for this task. 
The highest agreement ($>58\%$) is observed for `specificity' and `relevance'. The lowest agreement of $31\%$ is seen for the `novelty' feature, likely due to its inherent subjectivity. GPT-4 chose `informativeness' as the key interestingness indicator, in contrast to the `novelty' feature chosen by human annotators in Section~\ref{sec:factdataset}. 

\paragraph{Interesting Facts and User Engagement:}
We first evaluate the quality of the interesting facts in the dataset we created in Section~\ref{sec:factdataset}, and their effect on the user engagement with a CTA. 
We provide crowd workers with 200 sampled facts,\footnote{Average binary inter-annotator agreement = $0.68$} and ask them if the fact is (i) interesting to them; and (ii) likely to engage users interacting with a CTA.  
In this experiment, we do not provide our feature schema from Section~\ref{sec:schema} to the annotators, letting the definition of `interestingness' remain subjective and open to interpretation.  
$89\%$ of the input facts were chosen as both interesting and engaging to CTA users, reinforcing the effectiveness of our feature schema in defining interestingness, and the utility of our interesting fact dataset.

\subsection{Conversation Quality Evaluation}
\label{sec:human-eval}

We randomly select $50$ user-CTA cooking conversations, having $63$ turns with interesting facts. %
We ask human judges if the facts within the conversations might bore, frustrate or distract users from their original goal. The judges negatively labeled $22\%$ of the turns with facts, possibly because those facts were not appealing enough to justify the additional fact-related turns (examples in Appendix~\ref{sec:appendix1}).

\paragraph{Interesting Fact Placement in Conversations:}
We compare different strategies of inserting facts during a conversation between users and CTAs. We select 20 seed conversations, and present the following variants to crowd workers:\footnote{Average binary inter-annotator agreement = $0.60$} (i) showing interesting facts only during search, before the user selects a task; (ii) showing interesting facts only during task execution; (iii) always showing facts after seeking user approval; (iv) always showing facts without seeking user approval; and (v) our proposed hybrid fact placement approach in Section~\ref{sec:design}. 
In $65\%$ of cases, human judges preferred hybrid conversations where facts were shown both before task selection and during execution. 
Asking user permission before showing a fact was preferred $60\%$ of the time. This further validates the UX design choices we made in Section~\ref{sec:design}.

\subsection{Live Testing}
\label{sec:livedeploy}

\paragraph{Setup:}

We test our proposed interesting facts feature experience within a CTA of a leading commercial voice assistant, for live interaction and use by thousands of users in the US. 
Our CTA was built on top of the base Alexa Prize infrastructure~\cite{agichtein2023advancing}. 
The design, implementation and evaluation details of our final set up are described in Sections~\ref{sec:factdataset},~\ref{sec:design},~\ref{sec:automatic-eval} and~\ref{sec:human-eval}. 
We study two versions of the CTA, i.e., with and without the interesting facts feature, via A/B testing over a period of 6 weeks. We obtain explicit real-user feedback on our system in the following ways: (i) we ask users if they liked the interesting facts shown to them; (ii) users rate their experience conversing with the CTA with a score from 1 to 5; and (iii) users provide additional feedback via a free-form text field at the end of the conversation.

\paragraph{User Satisfaction and Engagement:}

We analyzed $450$ user-CTA conversations, of which $300$ belonged to the control group, and $150$ were exposed to our interesting facts UX. 
Users stated that they liked about $66\%$ of the interesting facts shown to them during the conversation. The average satisfaction rating provided by users to their conversations grew by $40\%$ (\textit{p}-value < $0.05$ with a \textit{t}-test) over the set up without the interesting facts. 

The average conversation length between users and the CTA grew by $37\%$ (\textit{p}-value < $0.05$ with a \textit{t}-test).
This is a favorable outcome overall, because while longer conversations can increase the amount of time spent by users to compete their task, shorter conversations are also likely to indicate task abandonment or user disinterest in continuing the conversation. Recall that the goal of our work is to maximize users' engagement with the CTA, since this can make our CTA more appealing for further adoption and usage (Sections~\ref{sec:intro} and~\ref{sec:related}). 
In fact, we noted a rise in the number of users both starting a task, and continuing their guided task interaction, after seeing an interesting fact. We also observed a $5\%$ increase in the task completion rate (i.e, the users completing all execution steps of their chosen task), and a $5\%$ increase in the repeat users interacting with the CTA after testing this feature. 

Finally, we examined users' next turn responses to the provided interesting facts, excluding those who explicitly mentioned liking the facts. We did not find any users abruptly ending their conversations after being presented with an interesting fact, or interrupting the CTA via voice as it spoke the fact. Manual inspection showed a positive or neutral sentiment in $>70\%$ of these user responses.  

\section{Discussion}
\label{sec:discussion}

\paragraph{Scope and Generalizability:}
The goals of our work are two-fold. The first is the generation of factually accurate interesting facts that are novel, specific, concise, informative and relevant to the task at hand. 
The second, more complex goal of our work is strategically injecting interesting facts into a user's conversation with a CTA, and involves multiple steps (Section~\ref{sec:design}). 
In this paper, we empirically evaluate our proposed approach only on the domain of cooking. However, we believe that the various steps and considerations involved in our framework, namely, our theoretically backed interestingness feature schema, fact curation and cleaning, and conversational presentation techniques can serve as an inspiration or a useful starting point for other applicable conversational task domains to consider, evaluate and expand upon. 

\paragraph{Scalability:}
Though potentially applicable across domains where interesting facts are available, our interesting fact dataset creation approach presented in Section~\ref{sec:factdataset} may not scale due to the manual annotation effort involved. We investigated the use of LLMs to annotate interesting facts in Section~\ref{sec:automatic-eval}. 
We next explore the automatic generation of interesting facts using an LLM (\texttt{gpt-3.5-turbo}). 
We sample $50$ entities from set $\mathcal{E}$ (Section~\ref{sec:factdataset}), and prompt GPT-3.5 to generate $50$ relevant and interesting facts %
as per our schema in Section~\ref{sec:schema}. 
The facts generated by GPT-3.5 are suitable for a user conversing with a CTA. owever, we found that $16\%$ of them are either factually inaccurate, or of questionable accuracy with no supporting evidence available. %
For instance, the fact: \textit{The Can Opener Museum located in Cincinnati, Ohio, celebrates the history and evolution of can openers} is  completely fabricated -- such a museum does not exist.

Therefore, directly showing unverified LLM-generated facts to users in an online manner %
can severely impair users' experience and trust in a commercial CTA that aims to %
be a reliably and knowledgeably guide users through costly and lengthy tasks. %
Online fact generation by an LLM during a conversation also increases the CTA's response latency. One solution, which we use in this work, involves offline correction and retrieval of facts before presenting to users. Another solution could be to use feedback dataset as demonstration data to tune an LLM or an explicit dialogue policy.

Using LLMs in commercial settings also poses a practical challenge, as legal requirements mandate appropriate crediting of original information sources. %
When prompted for the original sources it uses to generate facts, GPT-3.5 responds: \textit{I'm unable to directly access external websites or provide specific sources for facts as my knowledge}. We will thus need to provide the LLM a retrieval or API-based search component to get these sources automatically, which adds a risk of hallucination. 
We leave the end-to-end automation of our approach with one or more LLMs for future work.

\section{Conclusions}
\label{sec:conclusions}
\vspace{-0.2cm}

Drawing on psychological theories on human interest, this work defines what users might find interesting during guided task-based conversations with CTAs, and empirically demonstrates the value of engaging and satisfying users with appealing external knowledge. Engaged users are, in fact, more likely to complete tasks and leverage the power of CTAs for complex task assistance. 
To achieve this goal, we defined a semi-automatic framework to build a dataset of $1,379$ contextual and interesting facts for the cooking domain, and investigated different options to effectively inject the facts into user-CTA conversations. 
Offline human evaluations and live testing on a multi-modal voice assistant showed significant gains in user satisfaction and engagement in a CTA enriched with our interesting facts experience.  %

\section*{Limitations and Future Work}

Our work did not consider the following aspects, which we leave for future work.

\paragraph{Domain Generalization:} Based on the theoretical backing of our framework, we fully expect it to generalize across multiple task domains feasible for a CTA. However, we would like to verify this empirically, by generating interesting facts for other task assistant domains such as home improvement, inserting those facts into user-CTA conversations, and addressing any potential domain-specific challenges that may occur, e.g. recognizing and linking complex named entities \cite{malmasi-etal-2022-semeval}. 

\paragraph{Fact Presentation: }
Our current approach only considers the presentation of interesting facts in the form of simple text. Our multi-modal CTA currently shows the entire fact text on the screen, and reads out the entire fact by voice. We seek to explore varied modes of presentation combining both voice and the display screen, such as diverse text formats, catchy summary phrases, or visual aids to accompany the facts. 

\paragraph{Utilization of LLMs:}
We plan to reduce the number of steps and manual effort involved in our approach to increase its scalability and generalizability, by employing one or more LLMs to automate the process in an end-to-end fashion. This involves using a retrieval augmented LLM with explicit hallucination reduction techniques to generate factually accurate and interesting facts based on our feature schema. We will then use the insights we discovered in this work as well as any existing UX specifications or constraints, to either prompt or fine-tune an LLM to decide when, where and how best to naturally incorporate interesting facts into task-specific conversations.

Furthermore, though we explored the ability of LLMs to annotate as well as generate interesting facts, we did not directly present the factually accurate interesting facts generated by LLMs to real-world users interacting with our CTA. 
Given the disagreement on interestingness between human and GPT-4's annotations in Section~\ref{sec:human-eval}, we would like to investigate the user engagement and user satisfaction obtained when users are presented facts generated and selected by GPT-4, and how this differs compared to the interesting facts preferred by humans annotators.

\section*{Acknowledgements}
We would like to express our sincere gratitude to Zhiyu Chen, Jason Choi, Saar Kuzi as well as the anonymous reviewers for their insightful feedback and comments on our work.

\bibliography{references}

\newpage

\appendix

\onecolumn

\section*{Appendix}

\section{Examples of Conversation Turns with Interesting Facts}
\label{sec:appendix1}

Table~\ref{tab:examples} shows some examples of conversation turns containing interesting facts, and their labels annotated by human annotators. 

\begin{table*}[h]
\small
\centering
\begin{tabular}{l|c|c}
\hline
\textbf{Entity} & \textbf{Conversation turn} & \textbf{Human annotation} \\
\hline
Ingredient: Sweet potato & \begin{tabular}[x]{@{}c@{}}Did you know that the vibrant colors of sweet potatoes\\can be extracted and used as a natural dye for fabrics? From facts.net.\end{tabular} & Interesting \\ \hline
\makecell{Recipe: Smoked Salmon \\ Crepes} & \begin{tabular}[x]{@{}c@{}}According to facts.net, crepe-making is \\ an art form in Japan.\end{tabular} & \makecell{Boring, frustrating \\ or distracting} \\ \hline
Ingredient: Sausage  & \begin{tabular}[x]{@{}c@{}}Here's something interesting from thefactsite.com: \\ Sausages play a key role in Australian politics!\end{tabular} & Interesting \\ \hline
Tool: Chopsticks & \begin{tabular}[x]{@{}c@{}}According to Wikipedia, the first chopsticks were used not as \\ eating utensils but for cooking, stirring and serving.\end{tabular} & \makecell{Boring, frustrating \\ or distracting} \\ \hline
Recipe: Mooncake & \begin{tabular}[x]{@{}c@{}}According to monoandco.com, half-baked mooncakes must be taken out \\ and cooled for 15 minutes before continuing baking.\end{tabular} & \makecell{Interesting} \\ \hline
Ingredient: Candy & \begin{tabular}[x]{@{}c@{}}From eatthis.com, weirdly enough, cotton candy was invented \\ in 1897 by a dentist.\end{tabular} & \makecell{Interesting} \\ \hline
Recipe: Chocolate cake & \begin{tabular}[x]{@{}c@{}}According to facts.net, the secret to a moist cake is to \\ use ingredients like buttermilk, sour cream, or yogurt, which add moisture.\end{tabular} & \makecell{Interesting} \\ \hline
Ingredient: Pumpkin spice & \begin{tabular}[x]{@{}c@{}}Here's an interesting fact from facts.net: pumpkin spice does not \\ actually contain pumpkins.\end{tabular} & \makecell{Interesting} \\ \hline
Tool: Baking tin & \begin{tabular}[x]{@{}c@{}}From organicfacts.net, aluminium is considered the best material for \\ a baking tin as it allows for a quick transfer of heat.\end{tabular} & \makecell{Boring, frustrating \\ or distracting} \\ \hline
Ingredient: Baking soda & \begin{tabular}[x]{@{}c@{}}According to tasty.co, baking soda must be replaced every month, \\ otherwise a bit more than the recipe calls for can be added.\end{tabular} & \makecell{Interesting} \\ \hline
\end{tabular}%
\caption{Examples of conversation turns annotated for interestingness by human annotators.}
\label{tab:examples}
\end{table*}

\end{document}